\documentclass[letterpaper, 10 pt, conference]{ieeeconf}  
\IEEEoverridecommandlockouts                              

\usepackage{graphics} 
\usepackage{epsfig} 
\usepackage{stfloats}
\usepackage{times} 
\usepackage{amsmath} 
\usepackage{amssymb}  
\usepackage[hang]{subfigure} 
\usepackage{cite}
\usepackage{siunitx}
\usepackage{soul}
\usepackage{color}
\usepackage{xcolor}
\usepackage[ruled,linesnumbered]{algorithm2e}
\usepackage{caption}

\soulregister\ref7  
\soulregister\cite7 
\soulregister\eqref7
\soulregister\prettyref7

\usepackage{multirow}
\usepackage{algpseudocode}
\usepackage{makecell}
\usepackage{booktabs}
\usepackage{threeparttable}

\bstctlcite{IEEEexample:BSTcontrol}

\usepackage{authblk}

\usepackage{prettyref}
\newrefformat{fig}{Fig.~\ref{#1}}
\newrefformat{sec}{Sec.~\ref{#1}}
\newrefformat{tab}{Tab.~\ref{#1}}
\newrefformat{algo}{Algo.~\ref{#1}}
\newrefformat{eq}{(\ref{#1})}

\usepackage{color} 

\begin{document}

\title{\LARGE \bf
Event Masked Autoencoder: Point-wise Action Recognition with Event-Based Cameras
}
\author{Jingkai Sun$^{1,2,*}$, Qiang Zhang$^{1,2,*}$, Jiaxu Wang$^{1}$, Jiahang Cao$^{1}$, Hao Cheng$^{1}$, Renjing Xu}
\affil[$^1$]{Hong Kong University of Science and Technology (Guangzhou)}
\affil[$^2$]{Beijing Innovation Center of Humanoid Robotics Co. Ltd.}
\affil[$*$]{equal contributions}

\maketitle
\thispagestyle{empty}
\pagestyle{empty}

\begin{abstract}
Dynamic vision sensors (DVS) are bio-inspired devices that capture visual information in the form of asynchronous events, which encode changes in pixel intensity with high temporal resolution and low latency. These events provide rich motion cues that can be exploited for various computer vision tasks, such as action recognition. However, most existing DVS-based action recognition methods lose temporal information during data transformation or suffer from noise and outliers caused by sensor imperfections or environmental factors. To address these challenges, we propose a novel framework that preserves and exploits the spatiotemporal structure of event data for action recognition. Our framework consists of two main components: 1) a point-wise event masked autoencoder (MA
E) that learns a compact and discriminative representation of event patches by reconstructing them from masked raw event camera points data; 2) an improved event points patch generation algorithm that leverages an event data inlier model and point-wise data augmentation techniques to enhance the quality and diversity of event points patches. To the best of our knowledge, our approach introduces the pre-train method into event camera raw points data for the first time, and we propose a novel event points patch embedding to utilize transformer-based models on event cameras. 
\end{abstract}

\section{Introduction}
\label{sec:intro}
DVS are bio-inspired vision sensors that mimic the functioning of the human retina\cite{gallego2020event}. They capture visual information in the form of asynchronous events that encode changes in luminance in a scene, rather than capturing images at a fixed frame rate as conventional cameras do. These events are generated at a very high temporal resolution and have been shown to provide high-quality information about motion and temporal changes, which are suitable for spiking neural networks~\cite{jiang2023fully}. Action recognition and event cameras are important in computer vision with many applications, such as reconstruction~\cite{wang2024evggs}, robotics, and human-computer interaction\cite{kong2022human}. Some DVS-based action recognition models\cite{wang2020st,wang2019space,plizzari2022e2} use the spatiotemporal method to recognize actions from the spatiotemporal patterns of DVS events, rather than relying on explicit feature extraction and classification steps as in traditional frame-based approaches. The event data stream consists of a number of events that include a pixel position, a timestamp, and a polarity indicating whether the change is an increase or decrease in luminance. Traditional action recognition methods rely on RGB videos or depth maps, which are not always suitable for fast and dynamic actions. Event-based cameras are an alternative to traditional cameras due to their high temporal resolution and low latency, and they can capture fast and dynamic actions more accurately. However, handling continuous streams of event data in real-time is a challenge and requires efficient data compression, representation, and recognition algorithms.


A common method for converting event data into 2D images is time-surface\cite{manderscheid2019speed}, which is a temporal representation of the events captured by the camera. 
The event data represented by time-surface can then be fed into popular image networks such as convolutional neural networks\cite{amir2017low} or vision transformers\cite{sabater2022event}. However, this frame-based method loses the details of temporal information when slicing the data into images. This results in the inability of this method to take into account the rapid changes and temporal relationships of objects in continuous time. 

\begin{figure}[t]
    \centering
    \includegraphics[scale=0.38]{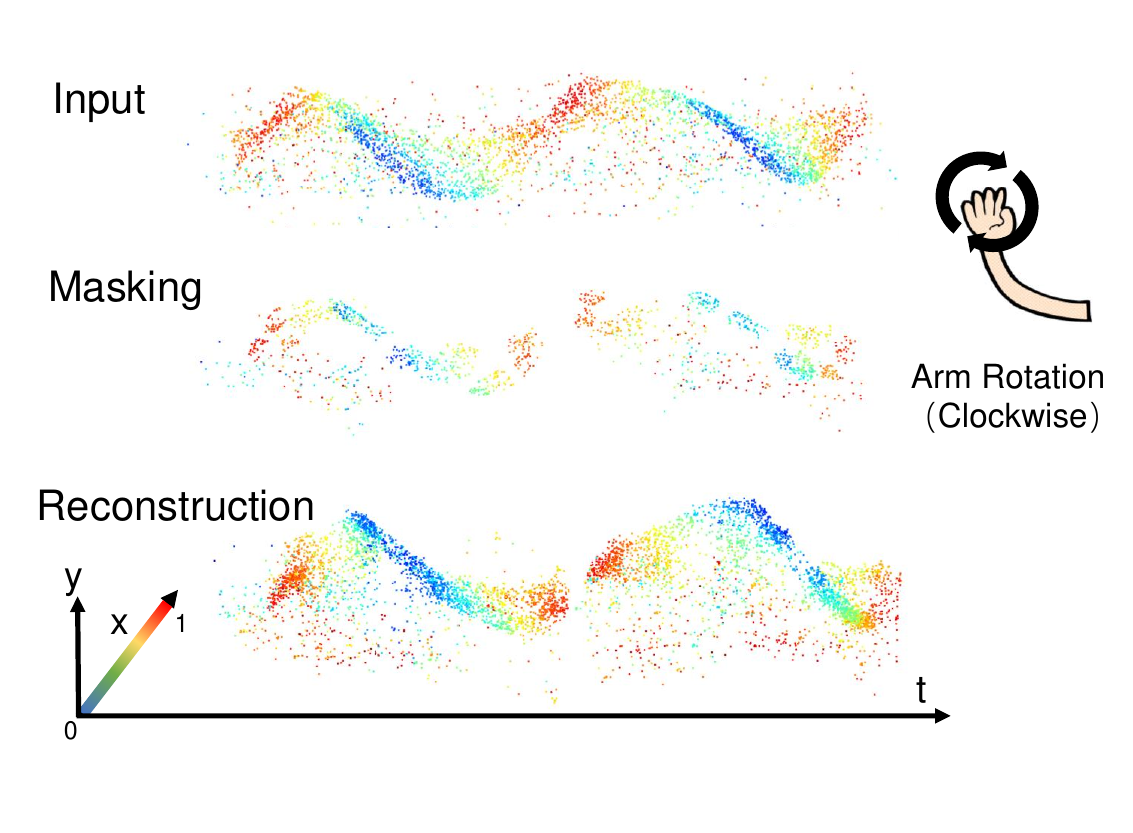}
    \vspace{-4mm}
    \caption{Reconstruction result of right arm rotation clockwise on DVS128-Gesture\cite{amir2017low}. We present the reconstruction results of a right arm rotation performed clockwise, using the DVS128-Gesture dataset. Specifically, we show the input data, the event data stream obtained after applying the mask, and the resulting reconstruction. In this example, we set the masking ratio to 80$\%$. Compared with the ground truth data, our reconstruction results are denser and less noisy. To visualize the results, we concatenate the 3 event data streams in time. The colors from blue to red indicate the magnitude of the values from 0 to 1 on the x-axis.}
    \vspace{-6mm}
    \label{Fig1}
\end{figure}

To address the three problems mentioned above, we adopt Point Masked AutoEncoder\cite{pang2022masked} as our base structure for event action recognition. MAE is a self-supervised learning neural network architecture that trains itself to reconstruct input data that has been partially masked. The masking process randomly removes some input tokens from the encoder's input during the forward pass, forcing the network to learn a compact representation of the data that can handle the missing information. However, the Farthest Point Sampling (FPS) algorithm is used to process the point cloud. It is not suitable for the event data streams. Because the farthest point describes the shape of the point cloud, it has better coverage of the entire point set. However, this does not work in event data for two reasons. First, unlike the point cloud, the Z-axis strictly describes the distance between each point, and the time axis does not describe the shape of the real object when the distance is calculated by X, Y, and time. Second, compared to point clouds, which are less affected by noise, event data is usually more heavily affected by noise\cite{stoffregen2020reducing} so FPS will sample more noisy points, which degrade the performance of MAE. Inspired by \cite{nagata2022self}, we improve the event patches generation algorithm. We utilize the event inlier model to choose suitable event patches. As shown in Figure \ref{Fig1}, the masked event data stream can be reconstructed easily by our method. On the basis of maintaining the original geometric shape, our method can also reduce part of the noise of the event data. For information loss caused by the sampling method, we use point-wise data augmentation called point resampling\cite{yu2022point} to improve our model.


We present a novel method that offers the following original and significant contributions:

1) We propose Event Masked Autoencoder, a novel method that treats event data streams as point clouds and applies masked modeling for the first time. Our method outperforms the state-of-the-art on several public benchmarks. We demonstrate that masked modeling methods are effective for event stream data. Moreover, we envision the possibility of using a unified backbone for the multi-modality fusion of event data, point clouds, and images in future work.

2) In this paper, we present a novel method of directly grouping event raw point data into patches and demonstrate its effectiveness through experiments. Our method can preserve the temporal and spatial information of event data and improve the performance of downstream tasks such as action recognition and reconstruction.

\begin{figure}[t]
    \centering
    \includegraphics[scale=0.24]{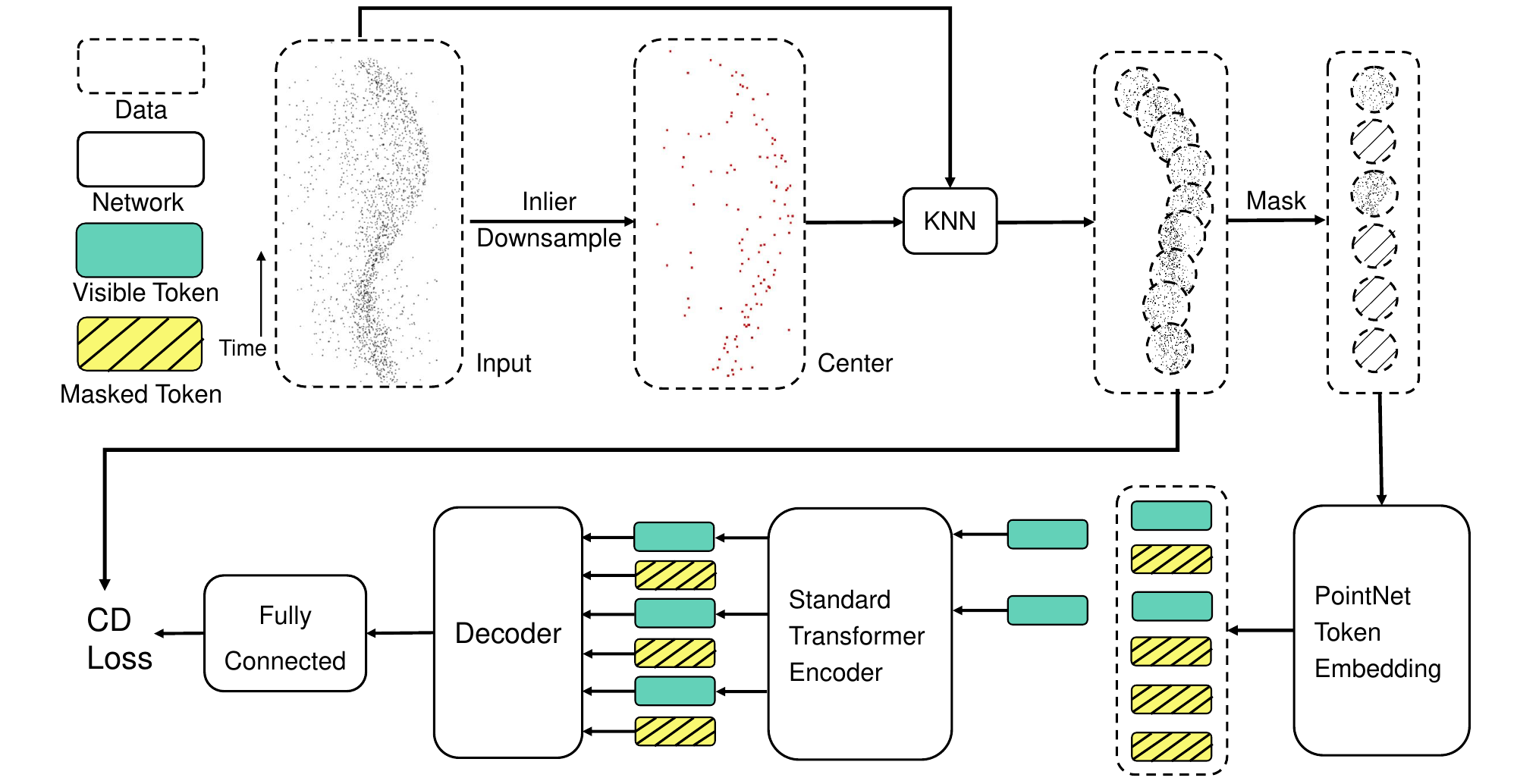}
    \caption{Pipeline of our Event Masked Autoencoder. In the first half, we present the event patch generation, mask, and embedding process. In the second half, we describe the pre-training process. Due to the asymmetric structure of the encoder and decoder, we differentiate them by their schematic sizes. During the encoding process, only visible tokens are fed into the network, while in the decoding process, masked tokens are added for reconstruction purposes.}
    \label{Fig2}
\end{figure}
\begin{figure}[t]
    \centering
    \includegraphics[scale=0.33]{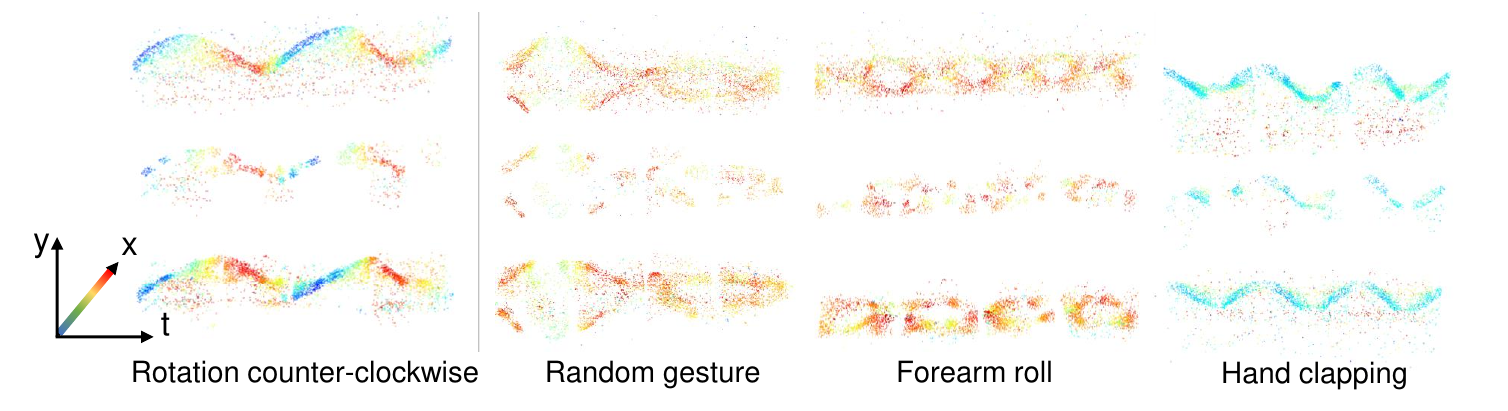}
\caption{Reconstruction examples on DVS128-Gesture test set. The colors from blue to red indicate the magnitude of the values from 0 to 1 on the x-axis(normalized).}
    \label{Fig3}
    \vspace{-4mm}
\end{figure}
\section{Method}
\subsection{Event Data Processing}
To convert $e_i$ into 3D point-wise data, we convert $t_i$ directly into $z_i$ and divide the event set into positive set $e_p = \{e_i | p_i=1,i\in\{0,\cdots, N\}\}$ and negative set $e_n = \{e_i | p_i=0,i\in\{0,\cdots, N\}\}$, where $N$ is the number of event in this event set. And we normalize the data with $t=\frac{t_n-t_0}{t_{max} - t_0}$, where $t_{max}$ is the max timestamp in this event set and $t_n$ is the timestamp of the $n_{st}$ event. Then we sample the event data stream by sliding window which normalizes the sampling range of the event set. 
According to previous research \cite{pang2022masked}, 1024 points are more suitable for Point-MAE based structure. However, in some sliding windows, the number of points is much greater than 1024. This caused the directly random sample will lose information. 
\subsection{Event Patch Generation For Masked Autoencoder }
In contrast to images, the event stream is composed of unordered points. Dividing contiguous neighboring pixels into patches is not available. Inspired by the Point-MAE\cite{pang2022masked}, we utilize the base masked autoencoder structure and introduce the event inlier model to improve generating the event patches.

\par As with point patches in Point-MAE, a single event patch is also composed of the center and its neighbors. To remove noise from the event stream and to determine the selected available centers, we fit a plane to the events of the local neighbors. The basic idea is that events belonging to the same plane should be located on or near the same plane in 3D space. Thus, by fitting a plane to a group of nearby events,  it is possible to select the available centers of the patch instead of selecting noise that does not fit this structure. The process of denoising by plane fitting usually consists of the following steps. First, a point of the event set is randomly selected as the initial solution $C_I$, depending on the number of patches. Then group nearby events into a local neighbor which is calculated by the K-Nearest Neighborhood (KNN) algorithm based on their proximity in space and time.
\begin{equation}
    P = KNN(E,C_I)
\end{equation}
where $E_i$ is a batch of data, $P \in \mathbb{R}^{k \times m \times C}$ is the total patches, $k$ is a hyperparameter of the K-Nearest Neighborhood which describes the number of points in a group, $m$ is the number of groups we set.
Second, fit a plane to the events in the neighborhood using a least-squares estimation method. We calculate separately for positive events and negative events. For each group calculated by KNN, let $A=[x_i,y_i,1]_{i=1,\dots,n}\in\mathbb{R}^{n\times3}$ , $n$ is the number of events in a group, $B=[t_1,\dots,t_n]^T$ and the parameters of least-squares method is $D=[a,b,c]$. To fit a plane of events in the group, we assume the movement of the object is linear in $\delta{t}, \delta{t}\rightarrow0$. So the plane can be defined by $t_i = ax_i+by_i+c$. The parameters can be estimated by
\begin{equation}
\Hat{D}=(AA^T)^{-1}A^TB    
\end{equation}

To normalize the coordinates, we choose the center as the coordinate origin and calculate them by $\Delta{x}_i=x_c - x_i$, $\Delta{y}_i=y_c - y_i$, $\Delta{t}_i=t_c - t_i$, where $x_c, y_c, t_c$ are the 3-dimensional coordinates of the center. Then, the normalized plane is
\begin{equation}
    \Delta{t_i} = a\Delta{x_i}+b\Delta{y_i}+c
\end{equation}
Then we calculate the average error 
\begin{equation}
    er_j=\frac{\sum{(\Delta{t_i}-\Delta{\hat{t_i}})}}{n}, j=1,\dots,m
\end{equation}
where $m$ is the number of groups we set.
To filter out unavailable patch centers that its group doesn't lie on or near the plane, we set a threshold $H$ as a hyperparameter to trade off the training speed and denoising effect. If the $er_j < H$, we add this event to the list of centers of patches. Then, repeat the process until we have a sufficient number of centers.

\begin{table}[t]
\caption{Classification Accuracy in DVS128-Gesture. N/A means it is not available at the source reference.}
\begin{center}
\begin{tabular}{l|cc}
\hline
\multicolumn{1}{c|}{\multirow{2}{*}{Model}} & \multicolumn{2}{c}{DVS128-Gesture}                   \\ \cline{2-3} 
\multicolumn{1}{c|}{}                       & \multicolumn{1}{c|}{10 Classes}     & 11 Classes     \\ \hline
Time Cascade\cite{amir2017low}                                         & \multicolumn{1}{c|}{96.49}          & 94.59          \\
PointNet++\cite{wang2019space}                                  & \multicolumn{1}{c|}{97.08}          & 95.32          \\
TORE\cite{baldwin2022time}                                        & \multicolumn{1}{c|}{N/A}            & 96.2           \\
EvT\cite{sabater2022event}                                         & \multicolumn{1}{c|}{98.4}           & 96.2           \\
CNN+LSTM\cite{innocenti2021temporal}                                    & \multicolumn{1}{c|}{97.5}           & 97.53          \\
\textbf{Our}                                & \multicolumn{1}{c|}{\textbf{98.54}} & \textbf{97.75} \\ \hline
\end{tabular}
\end{center}
\label{Table1}
\end{table}
\begin{table}[t]
\caption{Classification Accuracy in SL-Animals-DVS. N/A means it is not available at the source reference.}
\begin{center}
\begin{tabular}{l|cc}
\hline
\multicolumn{1}{c|}{\multirow{2}{*}{Model}} & \multicolumn{2}{c}{SL-Animals-DVS}                   \\ \cline{2-3} 
\multicolumn{1}{c|}{}                       & \multicolumn{1}{c|}{S3}         & S4         \\ \hline
SLAYER\cite{vasudevan2020introduction}                                      & \multicolumn{1}{c|}{78.03}          & 60.09          \\
STBP\cite{vasudevan2020introduction}                                       & \multicolumn{1}{c|}{71.45}          & 56.20          \\
TORE\cite{baldwin2022time}                                        & \multicolumn{1}{c|}{N/A}            & 85.1           \\
EvT\cite{sabater2022event}                                         & \multicolumn{1}{c|}{87.45}          & \textbf{88.12}          \\
DECOLLE\cite{kaiser2020synaptic}                                     & \multicolumn{1}{c|}{77.6}           & 70.6           \\
\textbf{Our}                                & \multicolumn{1}{c|}{\textbf{88.23}} & 87.46 \\ \hline
\end{tabular}
\end{center}
\label{Table2}
\vspace{-6mm}
\end{table}
\subsection{Event Masked Autoencoder Structure}

Following the masked and embedding method in Point-MAE, we mask the event patches with a high radio randomly. To apply patch embedding in an event-masked autoencoder.
each patch is first flattened into a vector and then concatenated together to form a sequence of vectors. This sequence is then fed into the encoder of the autoencoder, which maps the input sequence to a lower-dimensional latent representation. 
\par The Figure \ref{Fig2} illustrates the overall pipeline of our method.  In the process of the encoder, we only feed the visible token without the masked part. In our approach, we utilize PointNet to extract event features as a point cloud for the embedding of each masked event patch rather than linear projection like ViT\cite{liu2021swin}. 
\begin{equation}
T_v=PointNet(P_v)
\end{equation}

where $T_v \in \mathbb{R}^{(1-\alpha)\times m \times C}$ are the visible tokens, $P_v \in \mathbb{R}^{(1-\alpha)\times m \times k \times 3}$ are the visible patches, $\alpha \in (0,1)$ is the masked radio.
\par Then, the output of decoder $D$ is fed into a reconstruction model that is a fully connected (FC) layer. The reconstruction model is to project $D$ back to vectors which are the reconstruction result of the masked event patches. And we reshape the result to the dimensions as same as the masked event patches. The final result $P^{pre}_m \in \mathbb{R}^{\alpha \times n\times k \times3}$ and ground truth $ P^{gt}_m \in \mathbb{R}^{\alpha \times n\times k \times3}$ that got by the masking process are used to compute Chamfer Distance as the pre-train loss. 

Following \cite{devlin2018bert}, we map the vectors of event patch centers to the embedding dimension with a Multilayer Perceptron (MLP) to feed the position information into the network.

\begin{table}[t]
\caption{Classification Accuracy in DVS Action.}
\begin{center}
\begin{tabular}{l|clll}
\hline
\multicolumn{1}{c|}{Method} & \multicolumn{4}{c}{DVS Action Acc(\%)} \\ \hline
Motion SNN \cite{liu2021event}                 & \multicolumn{4}{c}{78.1}               \\
HMAX SNN \cite{xiao2019event}                   & \multicolumn{4}{c}{55.0}               \\
EV-ACT\cite{gao2023action}                   & \multicolumn{4}{c}{92.1}               \\
ST-EVNet\cite{wang2020st}                    & \multicolumn{4}{c}{88.7}               \\
PointNet \cite{qi2017pointnet}                   & \multicolumn{4}{c}{75.1}               \\
\textbf{Our}                         & \multicolumn{4}{c}{\textbf{93.9}}               \\ \hline
\end{tabular}
\vspace{-5mm}
\end{center}
\label{TableAction}
\end{table}
\section{Experiments}
\label{sec:sdiscussion}
In this section, we introduce the experiments of our approach. This process is divided into two parts, pre-train and fine-tune for downstream classification tasks. For various types of data augmentation, we use the most useful method point resample to process the training dataset.

\subsection{Pre-train Process}
First, we use the model pre-trained by ShapeNet\cite{chang2015shapenet} which is a popular point cloud dataset. According to our experiments, although point cloud and event are two different types of data, the model pre-trained on ShapeNet is better than the model trained only on event data. This is because point clouds and events have different global features, but they have similarities in local geometric information. So the model pre-trained by ShapeNet is a great initial solution before we feed event data into it. Compared to training the model from scratch, pre-training the model with the ShapeNet dataset and further training it with the event camera dataset led to a decrease of $2.3\times10^{-4}$ in the loss function. For the further pre-train with DVS128-Gesture\cite{amir2017low}, we set the number of input events as a typical  $N=1024$. The number of event patches is $m = 64$. The number of points in each patch is $k=32$ for KNN. We both pre-train with DVS-Gesture and SL-Animals-DVS. For the sliding window, we set the size of the window as 0.5s and the size of the step as 0.25s. Figure \ref{Fig3} illustrates the visualized reconstruction results.

\begin{figure}[htbp]
    \centering
    \includegraphics[scale=0.26]{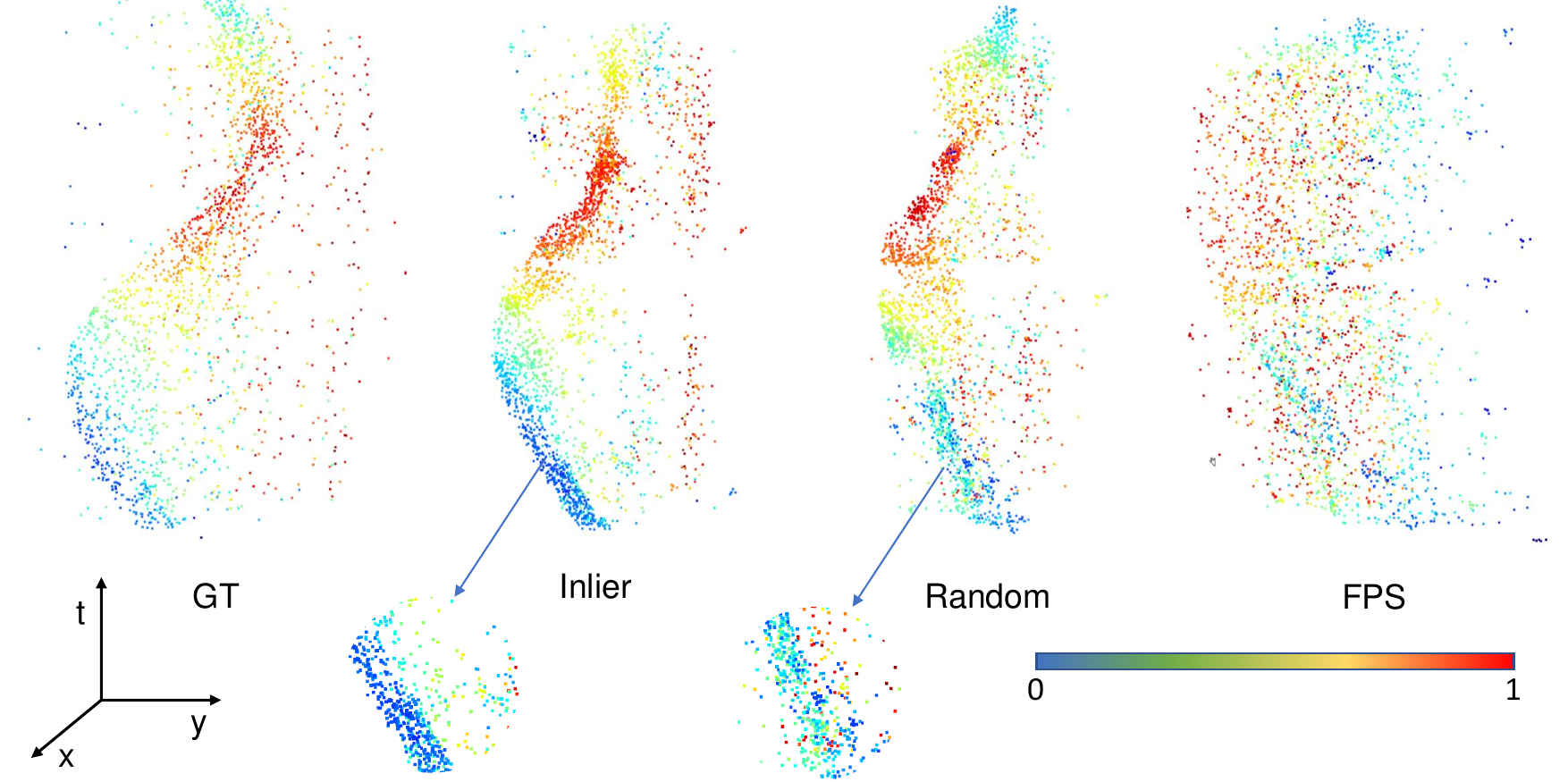}
    \caption{A comparison of three algorithms for generating event patches. We present the ground truth (leftmost), our approach (second from left), random sampling, and FPS (rightmost). Below the figure, we also provide some details about our approach and random sampling. From these details, we can evidently discern that our method has superior performance. The colors from blue to red indicate the magnitude of the values from 0 to 1 on the x-axis.
}
    \label{Fig4}
    \vspace{-7mm}
\end{figure}
\begin{table}[t]
\caption{Ablation study about three thresholds for our inlier method. }
\begin{center}
\begin{tabular}{l|c|c}
\hline
Threshold & Loss($\times1000$)           & Acc.(\%)       \\ \hline
0.35       & 1.283          & 95.92          \\
0.85      & \textbf{1.103} & \textbf{97.75} \\
1        & 1.213          & 96.83          \\ \hline
\end{tabular}
\end{center}
\label{Table4}
\vspace{-6mm}
\end{table} 
\begin{table}[t]
\caption{Ablation study about three methods for downsampling in event patch generation. }
\begin{center}
\begin{tabular}{l|cc}
\hline
\multicolumn{1}{c|}{\multirow{2}{*}{Sample Method}} & \multicolumn{2}{c}{Chamfer Distance($\times1000$)}                                                 \\ \cline{2-3} 
\multicolumn{1}{c|}{}                               & \multicolumn{1}{c|}{DVS128-Gesture}                  & SL-Animals-DVS                  \\ \hline
FPS                                                 & \multicolumn{1}{c|}{4.065}                           & 4.243                           \\
Random Sample                                       & \multicolumn{1}{c|}{1.225}                           & 1.253                           \\
Our proposed method                                 & \multicolumn{1}{c|}{\textbf{1.103}} & \textbf{1.161} \\ \hline
\end{tabular}
\end{center}
\label{Table3}
\vspace{-4mm}
\end{table} 
\begin{figure}[t]
    \centering
    \includegraphics[scale=0.30]{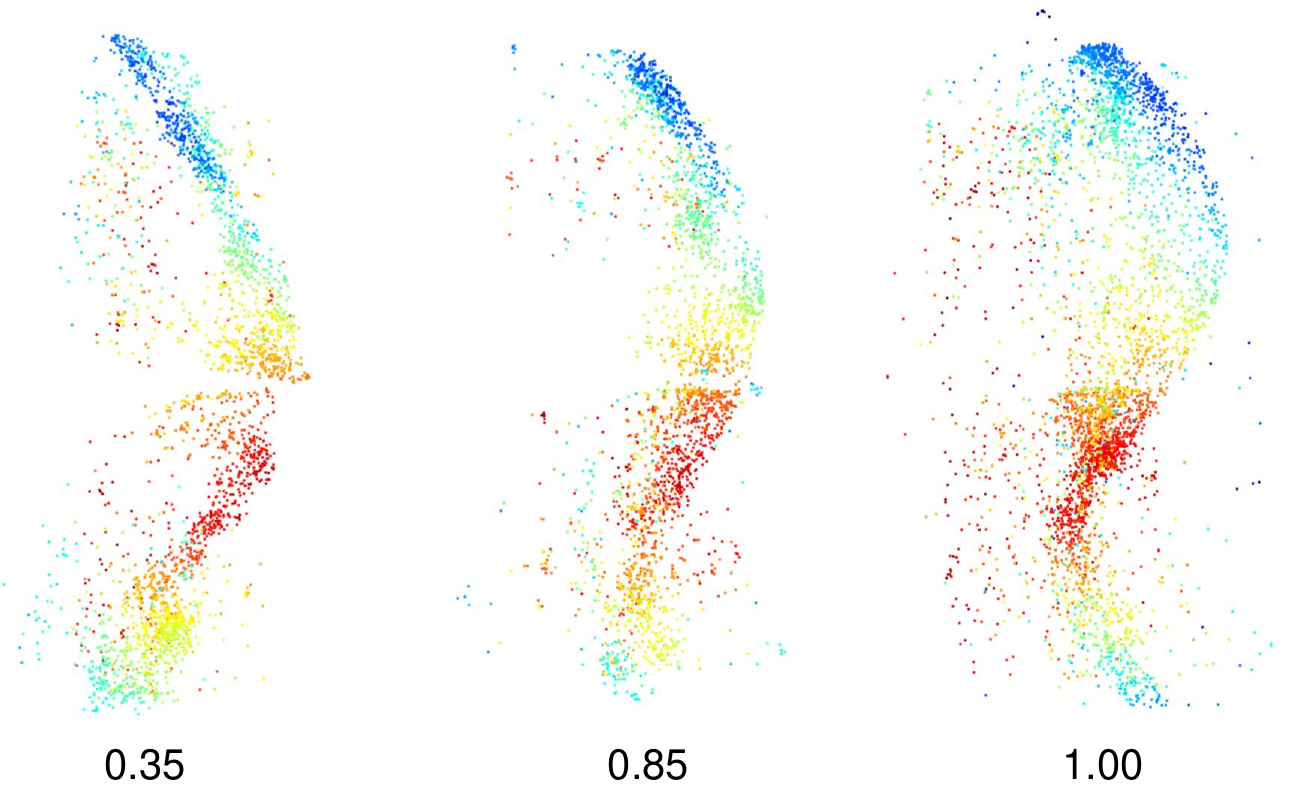}
\caption{We show three different thresholds. For visualization, we contact two event data streams. When the threshold is 0.35, the connection between the two event data streams is very sparse. When the threshold is 1, the event data stream has more noise.}
    \label{Fig5}
    \vspace{-7mm}
\end{figure}

\subsection{Downstream Classification Tasks}
SL-Animals-DVS\cite{vasudevan2021sl} is a dataset of sign language gestures performed by representing animals recorded with a DVS sensor. It has about 1100 samples performing 19 classes of sign language signs. We compare our approach with various methods both point-based and frame-based. The dataset is divided into 2 sets, S3 and S4. Set S3 was recorded with artificial lightning from a neon light. Set S4 was recorded under strong sunlight which caused more noise. The results show that our method has state-of-the-art performance in Set S3. It indicates that our approach is not good at dealing with the event data with more noise. Based on our observations, we hypothesize that the inefficacy of the event patches generation modal in removing noise is the underlying cause of the phenomenon.

DVS128 Gesture Dataset contains recordings of 10 or 11 (including random gestures) different hand gestures. DVS Action is a dataset smaller than DVS128 Gesture Dataset. So we do not fine-tune our model on this dataset. 

Table \ref{Table1} shows the top-performing online models evaluated on the DVS128-Gesture dataset. We obtain state-of-the-art performance in the DVS128-Gesture dataset both 10 classes and 11 classes. Table \ref{Table2} shows the top-performing models in SL-Animals-DVS Dataset. Our approach achieves state-of-the-art in S3 which is less noisy than S4. This indicates that our method is more robust to noise compared to the point-based method, and performs better with clean event data than the frame-based method. Table \ref{TableAction} shows the results in the DVS Action dataset.

\subsection{Ablation Study}
We evaluate the method using FPS, random sample, and our proposed method and present the results in a tabular format. Our findings suggest that FPS is a more suitable approach for point cloud processing. However, for event data, its performance is inferior to that of the random sample. We attribute this difference to the fact that FPS tends to collect a significant number of edge noise points, which can be mitigated to a certain extent by our proposed method. Figure \ref{Fig4} illustrates the comparison of three event patch generation algorithms. The FPS can't reconstruct the geometry of the event. Compared with the random sample, our method can be dense in detail and have less noise. The results of loss are shown in Table \ref{Table3}. Our approach is capable of replacing FPS for selecting event patches at the center with remarkable precision. Compared to FPS, our method can reduce the reconstruction loss during pre-training by almost fourfold, and can even enhance the results by one level through random sampling. The reconstruction loss and accuracy of downstream classification tasks on DVS128-Gesture are shown are Table \ref{Table4}. Based on our experiments, we found that selecting a threshold that is either too large or too small can result in poor model performance. If the threshold is too large, the sampling approach will be very similar to random sampling, which will not filter out noise effectively. On the other hand, if the threshold is too small, will lead to excessive aggregation of event data, resulting in a loss of information with a low density of boundary points. We visualize the reconstruction results with different thresholds in Figure \ref{Fig5}.

\bibliographystyle{IEEEtran}
\typeout{}
\bibliography{IEEEabrv,mybibfiles}
\end{document}